\documentclass{article}
\usepackage{arxiv}

\usepackage[utf8]{inputenc}

\usepackage[ngerman,english]{babel}

\usepackage[autostyle]{csquotes}
\usepackage[%
    backend=biber,%
    style=LNI,    %
    natbib=true   %
]{biblatex}%
\bibliography{references}
\AtEveryBibitem{%
  \ifentrytype{article}{%
  }{%
    \clearfield{doi}%
    \clearfield{issn}%
    \clearfield{url}%
    \clearfield{urldate}%
  }%
  \ifentrytype{inproceedings}{%
  }{%
    \clearfield{doi}%
    \clearfield{issn}%
    \clearfield{url}%
    \clearfield{urldate}%
  }%
}

\usepackage{url}

\usepackage{booktabs}

\usepackage{amssymb}
\usepackage{mathtools}
\usepackage{bm}
\usepackage{xspace}
\usepackage{microtype}
\usepackage{multicol}
\usepackage{booktabs}
\usepackage{nth}

\usepackage[nolist,nohyperlinks]{acronym}

\usepackage{multirow}
\usepackage{tikz}
\usepackage{pgfplots}
\usepgfplotslibrary{fillbetween}
\pgfplotsset{compat=newest}
\usetikzlibrary{patterns,
pgfplots.groupplots,fit,calc,positioning,shapes,spy}

\usepackage{tabularx}
\usepackage{booktabs}

\usepackage[breaklinks=true,colorlinks,bookmarks=false,citecolor=green!80!black,linkcolor=red!80!black,urlcolor=blue]{hyperref}
\usepackage{cleveref}

\usepackage{subcaption}
\usepackage{csquotes}
\usepackage{siunitx}
\DeclareSIUnit{\mio}{\text{mio.}}

\usepackage[inline]{enumitem}
\setlist*[enumerate]{label=(\arabic*)}

\newcommand{\onedot}{.\xspace}

\newcommand{\etal}[1]{#1~et~al\onedot}
\newcommand{\eg}{e.\,g.,\xspace}
\newcommand{\cf}{cf\onedot}
\newcommand{\ie}{i.\,e.,\xspace}

\newcommand{\gw}{\textsc{gw}\xspace}
\newcommand{\tw}{\textsc{tw}\xspace}
\newcommand{\longs}{{\fontencoding{TS1}\selectfont\char115}}

\newcommand{\map}{mAP\xspace}

\definecolor{faublue}{RGB}{0,51,102}
\definecolor{darkgreen}{rgb}{0,0.6,0}
\definecolor{bblue}{HTML}{4F81BD}
\definecolor{rred}{HTML}{C0504D}
\definecolor{ggreen}{HTML}{9BBB59}
\definecolor{ppurple}{HTML}{9F4C7C}

\begin{document}

\begin{acronym}
  \acro{cnn}[CNN]{Convolutional Neural Network}
\end{acronym}

\newcommand*{\email}[1]{\href{mailto:#1}{\urlstyle{same}\protect\nolinkurl{#1}}}
\renewcommand{\author}{\@dblarg\@@author}
\def\@@author[#1]#2{\gdef\@shortauthor{{\let\footnote\@gobble%
   \def\and{\unskip,\ }#1}}\gdef\@author{#2}}

\title{Proof of Concept: Automatic Type Recognition}

\author
{Vincent Christlein$^\star$, Nikolaus Weichselbaumer$^\dagger$, Saskia Limbach$^\ddagger$,
Mathias Seuret$^\star$
\\
$^{*}$Friedrich-Alexander-Universität Erlangen-Nürnberg, Pattern Recognition Lab, Germany
\\
{\tt\small vincent.christlein@fau.de, mathias.seuret@fau.de}
\\
$^\dagger$Johannes Gutenberg-Universität Mainz, 
Gutenberg-Institut für Weltliteratur und schriftorientierte Medien, 
\\Abt. Buchwissenschaft,
Germany
\\ 
{\tt\small weichsel@uni-mainz.de}
\\
$^\ddagger$ 
Università degli Studi di Milano,
Department of Economics, Management and Quantitative Methods,\\
Via Conservatorio 7, 20122 Milano, Italy\\
{\tt\small s.limbach@gmail.com}
}
\maketitle

\begin{abstract}
The type used to print an early modern book can give scholars valuable information about the time and place of its production as well as its producer.
Recognizing such type is currently done manually using both the character shapes of `M' or `Qu' and the size of the total type to look it up in a large reference work. This is a reliable method, but it is also slow and requires specific skills. 
We investigate the performance of type classification and type retrieval using a newly created dataset consisting of easy and difficult types used in early printed books.
For type classification, we rely on a deep Convolutional Neural Network (CNN) originally used for font-group classification while we use a common writer identification method for the retrieval case. 
We show that in both scenarios, easy types can be classified/retrieved with a high accuracy while difficult cases are indeed difficult. 

\end{abstract}

\section{Introduction}

Type recognition is one of the central methods of analytical bibliography \cite[p.~42–70]{schmitz2018}. 
It is used to date printed books and identify both the printer and the publication place. 
Type recognition is traditionally done manually by using a combination of the type size (measured over 20 lines) and the characteristic shape of the letters `M' or `Qu' to identify the correct type in the Typenrepertorium der Wiegendrucke (\tw), a reference work for all known incunabula types~\cite{haebler1905}.\footnote{The Typenrepertorium der Wiegendrucke is available online: \url{https://tw.staatsbibliothek-berlin.de}} 
This method is slow and requires specific skills. 
Most importantly, it relies on the existence of a reference work that is difficult and time-consuming to compile, making it next to impossible to use type recognition for material beyond the incunabula period.
For this approach, pattern recognition methods are extremely helpful as they increase the speed and ease of type recognition and thereby widen the scope of material that it can be applied to.
For book historians, using type recognition for books printed in the incunabula period and beyond helps to answer important questions.
In the early modern period, many books appeared without any indication about when and where the book was printed. 
This was often done when authors and printers feared political prosecution, but not limited to such cases.
Type recognition would enable us to identify the producers of these books. 
On top of that, we would gain a better understanding of the material used in a given print shop, which could tell us more about the economic background of the printer. 
This paper explores the effectiveness of existing pattern recognition methods from writer identification and font group recognition.

\section{Related Work}
We want to investigate type recognition in two different ways: classification and retrieval. 

\paragraph{Classification}
An early work for font classification~\cite{Wang15} – working on modern computer fonts –  built a large dataset of real and mostly synthetic images of about \num{5000} classes and \SI{2}{\mio} images. 
The proposed classifier achieves about \SI{80}{\percent} accuracy on the test set. 
Closer related are the competitions and datasets for cursive script type classification~\cite{Cloppet16,Cloppet17}, where the best method can differentiate between \num{12} script classes with an accuracy of about \SI{85}{\percent}~\cite{Christlein19ICDAR,Cloppet17}.
In contrast, font group classification~\cite{hip2019}, seems to be an easier task achieving accuracies of about \SI{98}{\percent}~\cite{dhd2020}. 
Note that we strive to classify types, which are much more challenging than font groups since the differences are often much smaller. 

\paragraph{Retrieval}
Type retrieval is closely related to other image retrieval tasks, such as writer retrieval for historical data. 
The current writer identification performance is well represented in the last image retrieval competitions~\cite{Fiel17ICDAR,Christlein19Comp}.
These competitions involved large datasets containing \num{3600}~\cite{Fiel17ICDAR} and \num{20000} test images~\cite{Christlein19Comp}. 
The accuracies vary widely (\SIrange{74}{97}{\percent}) depending on the data source and image quality.
The current state-of-the-art approach for historical writer identification is given by %
\etal{Lai}~\cite{Lai20}.
They propose ``pathlet'' features, which they combine with SIFT~\cite{Lowe04DIF} and encode it in a novel bagged version of VLAD encoding~\cite{Jegou12ALI}. 
They achieve about \SI{90}{\percent} Top-1 accuracy and outperform the previous
unsupervised deep learning-based approach by \etal{Christlein}~\cite{Christlein17ICDAR}. 
In this study, we evaluate the performance of a baseline writer identification method~\cite{Christlein19PHD} based on SIFT descriptors.

\section{Dataset} \label{sec:dataset}
\tw lists about \num{6000} different types used between 1450-1500.
Yet, it is impossible to use all of these types in this dataset. 
The sheer size of \tw already presents a challenge, but it is mainly its design and original purpose which prevents us from using all of it. 
\tw only lists the type used in a given book, but gives no indication if this type is the only type used in the book and – more importantly in this context – on which pages the type is used.  
Most books were printed with more than one type. Additional types were used in order to emphasize words, highlight headlines, etc. 
As of now the only way to select training data is to manually label every image. 
This made us limit our dataset to 8 examples which were selected to illustrate how easy and how difficult it can be to differentiate between various types.

\begin{figure}[t]
\begin{subfigure}[t]{\textwidth}
\centering
        \href{https://api.digitale-sammlungen.de/iiif/presentation/v2/bsb00041740/canvas/19/view}{\frame{\includegraphics[width=0.49\columnwidth]{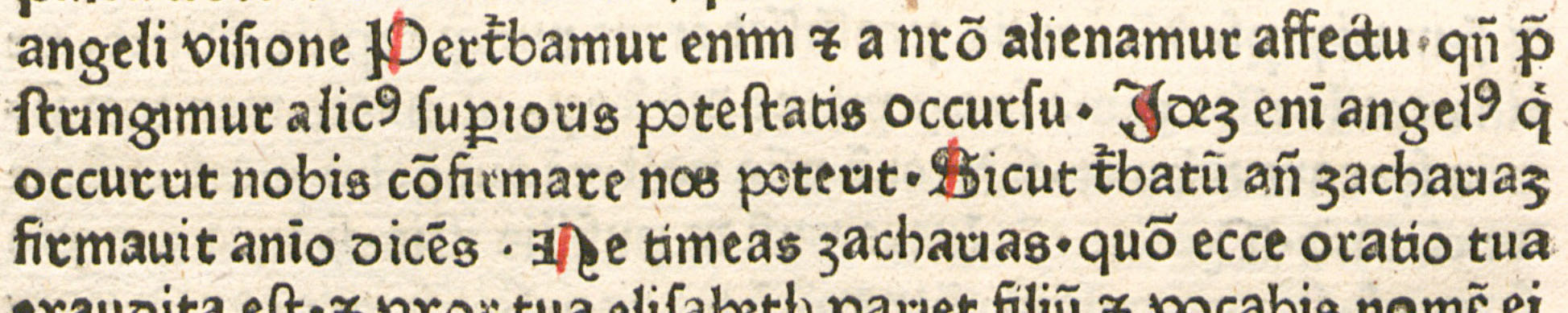}}}
            \hspace{1px}
            \href{https://bvpb.mcu.es/es/catalogo_imagenes/grupo.do?path=5017&posicion=28&presentacion=pagina&registrardownload=0}{\frame{\includegraphics[width=0.49\columnwidth]{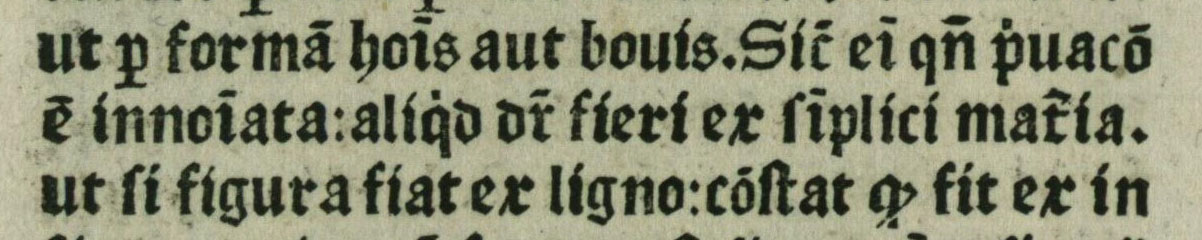}}}
            \par \vspace{4px}
            \href{https://gallica.bnf.fr/ark:/12148/btv1b86042919/f87.image}{\frame{\includegraphics[width=0.49\columnwidth]{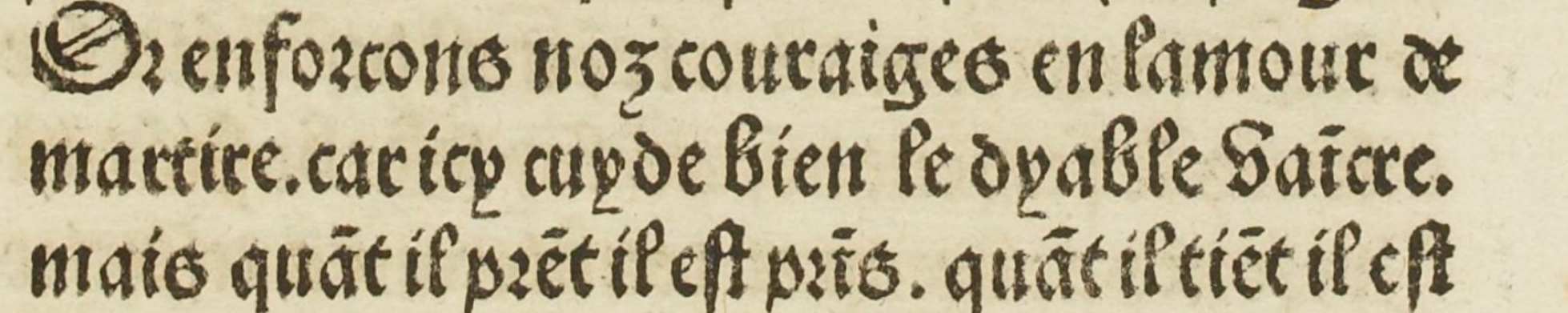}}}
            \hspace{1px}
            \href{http://dlib.gnm.de/item/N76/29}{\frame{\includegraphics[width=0.49\columnwidth]{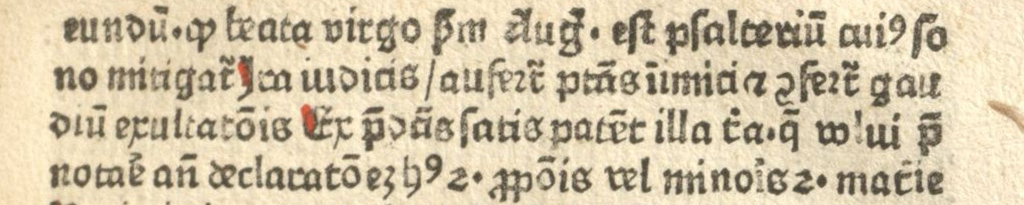}}}
        \caption {Easy cases – \href{https://tw.staatsbibliothek-berlin.de/ma00131}{\tw~ma00131} (top left), 
\href{https://tw.staatsbibliothek-berlin.de/ma00967}{\tw~ma00967} (top right), 
\href{https://tw.staatsbibliothek-berlin.de/ma02771}{\tw~ma02771} (bottom left),  
\href{https://tw.staatsbibliothek-berlin.de/ma04614}{\tw~ma04614} (bottom right).} 
        \label{fig:easy}
\end{subfigure}
\smallskip

\begin{subfigure}[t]{\textwidth}
\centering
        \href{https://daten.digitale-sammlungen.de/bsb00029982/image_9}{\frame{\includegraphics[width=0.49\columnwidth]{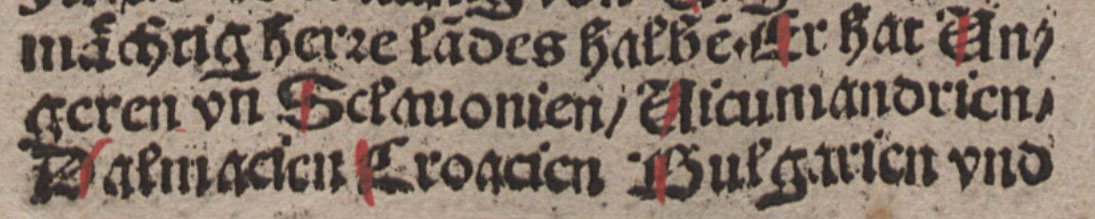}}}
        \hspace{1pt}
        \href{https://daten.digitale-sammlungen.de/bsb00033577/image_12}{\frame{\includegraphics[width=0.49\columnwidth]{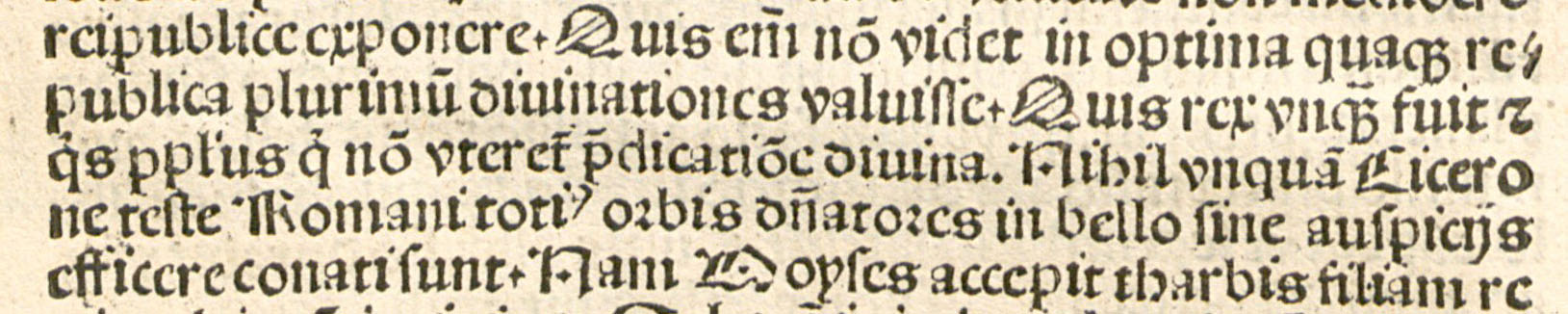}}}
    \caption{Difficult case 1 – \href{https://tw.staatsbibliothek-berlin.de/ma07487}{\tw~ma07487} (left) and \href{https://tw.staatsbibliothek-berlin.de/ma07488}{\tw~ma07488} (right).}
    \label{fig:difficult1}
    \end{subfigure}
 \smallskip   
    
    \begin{subfigure}[t]{\textwidth}
    \centering
        \href{https://daten.digitale-sammlungen.de/bsb00101728/image_1}{\frame{\includegraphics[width=0.49\columnwidth]{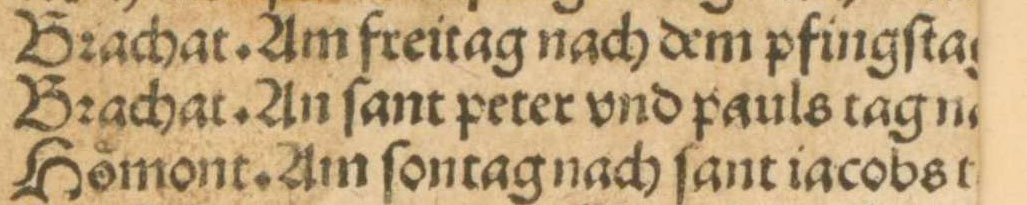}}}
        \hspace{1px}
        \href{https://daten.digitale-sammlungen.de/bsb00026982/image_31}{\frame{\includegraphics[width=0.49\columnwidth]{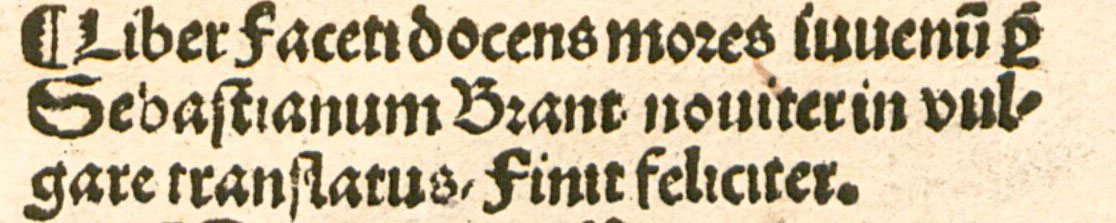}}}
    \caption{Difficult case 2 – \href{https://tw.staatsbibliothek-berlin.de/ma07721}{\tw~ma07721} (left) and  \href{https://tw.staatsbibliothek-berlin.de/ma07718}{\tw~ma07718} (right).}
    \label{fig:difficult2}
\end{subfigure}
\caption{Type dataset examples.}
\label{fig:type_examples}
\end{figure}

The easy examples, \cf \cref{fig:easy}, consist of types with very distinctive shapes. \href{https://tw.staatsbibliothek-berlin.de/ma00131}{\tw~ma00131} is a Rotunda used by the Augsburg printer Anton Sorg between 1475 and 1478 in at least 19 editions of which we have a total of 21 digital copies. The type is very recognizable because of its unique decorative upper case characters. 
Note, the red highlights were later added by a contemporary hand and not printed.

\href{https://tw.staatsbibliothek-berlin.de/ma00967}{\tw~ma00967} is a Textura used by an unknown printer from Salamanca from 1481 to 1490. While there is not as much surviving material for this type as for others (only 8 editions and 4 digital copies), the type is still easy to recognize for a human expert. 
The type is of rather poor quality and has a characteristic jagged look.  

\href{https://tw.staatsbibliothek-berlin.de/ma02771}{\tw~ma02771} is a Bastarda used by Jean du Pré in Lyon between 1489 and 1491. Relatively little material survives – \tw lists 8 editions and only 2 digital copies –, but the type is rather large at 119 mm over 20 lines and with its looped ascenders and flourished upper case characters it uses very complex shapes that are often easier to attribute correctly than the regular shapes of \eg a well-cut Textura. 

\href{https://tw.staatsbibliothek-berlin.de/ma04614}{\tw~ma04614} is a Textura with some unusual letter shapes used by Arnold ter Hoernen in Cologne from 1474 to 1482. It survives in 11 editions of which 5 copies are available as scans. The type is unusual in that most of its letter shapes follow the model of a Textura, but `\longs' and `f' have descenders and the lower-case `a' is of the cc-type, not the uncial type that is more common for Textura types.

To represent the other end of the spectrum we selected two pairs of types that we assumed to be particularly challenging.
\href{https://tw.staatsbibliothek-berlin.de/ma07487}{\tw~ma07487} and \href{https://tw.staatsbibliothek-berlin.de/ma07488}{\tw~ma07488} were both used by Bartholomäus Kistler in Straßburg, ma07487 from 1498 to 1499 (2 editions, 2 digital copies) and ma07488 from 1499 to 1501 (6 editions, 7 digital copies). 
The types share identical upper case Rotunda characters, but combine them with Bastarda lower case (ma07487) and Rotunda lower case (ma07488). 
Presumably, ma07487 was a temporary fix as the production of the lower case was not yet finished. 
This kind of combination appears fairly often and poses particular challenges as parts of the types are not similar, but completely identical. 

The second pair of difficult types, \href{https://tw.staatsbibliothek-berlin.de/ma07721}{\tw~ma07721} and \href{https://tw.staatsbibliothek-berlin.de/ma07718}{\tw~ma07718} were successively used by Johann Schäffler in Ulm from 1492 to 1494 (5 editions, 2 digital copies) and from 1496 to 1500 (22 editions, 11 digital copies). Both types are Upper Rhine Bastardas of very similar size and design. They do however not share any perfectly identical characters. \tw~ma07718 only replaced \tw~ma07721 after Schäfflers attempt to establish a print shop in Freising. It was apparently made to replace the older type~\cite[p.\,370–371]{Amelung1979}.  
These types can be told apart by a human expert, but only by methodical comparison. At a cursory examination they can easily be mistaken for one another. 

To create training data we used \tw and its sister catalogue the Gesamtkatalog der Wiegendrucke (\gw)~\cite{gw1925}.
For a given type we looked up all editions that are known to contain this type in \tw. Via \gw we searched for and downloaded all accessible scans of these editions. In the next step, we selected pages that contained only the respective type and in some cases cropped images in order to delete headlines, woodcuts and other irrelevant material. The dataset consists of 9083 labeled images and is publicly available.\footnote{\url{https://doi.org/10.5281/zenodo.3923638}}

\section{Classification}
\label{sec:classification}
As a first step to approach this problem, we train a classifier to recognize the selected types.
We use the same methodology as in~\cite{hip2019}.
The baseline method of this work consists of a \ac{cnn} to classify the text of document images as belonging to different font groups, such as Antiqua, Fraktur or Textura.

\subsection{Methodology}
We use a DenseNet-121~\cite{huang2017densely}, a deep \ac{cnn} with densely-connected blocks, to classify overlapping patches over the whole surface of the input image, and average the results.
We took the pre-trained one provided by Seuret~\textit{et al.}\footnote{\url{https://github.com/seuretm/ocrd_typegroups_classifier}}~\cite{hip2019}, and replaced its last layer by a new one with eight outputs.
As this network has been trained for font groups classification, we expect it to have learned features useful for type classification.
Note that we train only the last layer, leaving the other ones frozen.
In earlier tests, which we conducted, we realized that fine-tuning the whole \ac{cnn} makes it over-fit quickly. 

The training is done as follows:
First, we create a training set consisting of 5000 patches of 300$\times$300 pixels for each type, uniformly distributed over the available training images.
Then, during the training of the \ac{cnn}, we apply data augmentation on these patches and extract 224$\times$224 pixels center crops.
We used the following data augmentation strategies: random rotations between $\left]-15,15\right[$ degrees, shearing of an angle between $\left]-3,3\right[$, re-scaling by a factor in the range of $\left]0.9,1.1\right[$, color jittering (PyTorch settings: 0.7, 0.7, 0.3, 0.03). Additionally, we add JPG artifacts with quality factors in the range of $\left[2, 100\right[$, and a low binarization probability with Otsu (\SI{5}{\percent}) or Sauvola (\SI{2.5}{\percent}).

The network is trained for 10 epochs, with an initial learning rate of 0.0005, a weight decay of 0.00001, and a momentum of 0.9.
After each epoch, the learning rate is decreased by \SI{5}{\percent}.
Due to the small amount of images for some types, no validation set is used.

\subsection{Results}
\begin{table}[t]
	\centering
    \caption{Confusion matrix containing the classification results. To enhance readability, the diagonal is bold. Rows correspond to types, and columns to classification results. The first four rows depict the easy cases while the last four rows show the difficult cases.}
    \label{tab:classification}
    \begin{tabular}{lcccccccc}
        \toprule
        prediction $\rightarrow$ & \scriptsize ma00131 & \scriptsize ma00967 & \scriptsize ma02771 & \scriptsize ma04614 & \scriptsize ma07487 & \scriptsize ma07488 & \scriptsize ma07718 & \scriptsize ma07721 \\
        \midrule
           ma00131 & \textbf{160} &   0 &   0 &   0 &   0 &   0 &   0 &   0 \\
           ma00967 &   0 & \textbf{115} &   0 &   0 &   0 &   0 &   0 &   0 \\
           ma02771 &   0 &   0 & \textbf{711} &   0 &   0 &   0 &   0 &   0 \\
           ma04614 &   2 &   0 &   0 & \textbf{225} &   0 &   0 &   0 &   0 \\
           \midrule
           ma07487 &   1 &   0 &   0 &   0 &   \textbf{0} &   6 & 225 &   0 \\
           ma07488 &   0 &   0 &   0 &   0 &   0 & \textbf{124} &  11 &   0 \\
           ma07718 &   0 &   0 &   0 &   0 &   0 &   0 &   \textbf{2} &   0 \\
           ma07721 &   0 &   0 &   0 &   0 &   0 &   6 & 1012 &   \textbf{0} \\
        \bottomrule
    \end{tabular}
\end{table}

For the evaluation, we split the dataset into a training and a test set. The training set consists of 6472 samples while the test set contains 2600 document images. %
We made sure that there is no document overlap in the subsets to guarantee document-independent testing.

A confusion matrix presenting the classification results is shown in \cref{tab:classification}.
We can see that the system reaches an overall classification accuracy of \SI{51.4}{\percent} and an average accuracy of \SI{73.9}{\percent}, 
which already indicates that some classes are well recognized in contrast to others. 
The overall accuracy is significantly lower than the accuracy obtained on the training data (over \SI{80}{\percent} of the patches), which implies that we are running into over-fitting despite the rather aggressive augmentation approach.\footnote{Note that a major constraint is that the augmentation should not make a type look like another one.}

While the classification of the easy types was successful, several documents of the difficult types could not be detected properly. 
What makes the training for type ma07721 especially difficult is the fact that there is only one single training sample available, even when cropping a sufficient number of patches, the script variance might be too low for a reliable training. 
This suggests the use of a retrieval scenarios where the learning of a good embedding is in focus. 

\section{Retrieval}
\label{sec:retrieval}
In addition to classification, we test the retrieval scenario. 
That means, we want to retrieve the most similar types given a query image.

\subsection{Methodology}
We make use of the general writer identification framework by Christlein~\cite{Christlein19PHD}. 
It consists of a sampling step, where we evaluate two strategies: SIFT
keypoints~\cite{Lowe04DIF} computed at (a)~the original images, (b)~contours extracted by means of the well-known Canny edge detector~\cite{Canny}. 
For the latter approach, we set the two hysteresis thresholds automatically~\cite{Wahlberg15}. 

Afterwards, SIFT descriptors~\cite{Lowe04DIF} are computed at the keypoint locations. 
They are Dirichlet-normalized and PCA-whitened following Christlein~\cite{Christlein19PHD}.
Afterwards, the local descriptors are encoded using VLAD~\cite{Jegou12ALI} encoding using 100 clusters for the codebook. 
For improving the VLAD embedding~\cite{Christlein18DAS}, we employ Generalized Max-Pooling (GMP)~\cite{Murray16} with $\lambda=1000$ in combination with power normalization (power of 0.5) and $\ell^2$-normalization, \ie normalizing the global descriptor such that its norm equals one.
This process is repeated five times, the resulting global descriptors are concatenated and jointly PCA-whitened and dimensionality reduced to 6400 components and $\ell^2$-normalized again. 

Finally, an Exemplar-SVM (ESVM) is computed, which has shown to improve the writer identification results~\cite{Christlein17PR}. 
We use the ESVM as a feature transformation~\cite{Christlein17ICDAR},
\ie the $\ell^2$-normalized coefficients of the ESVM are used as new feature
descriptor.
These descriptors are then compared using the cosine distance, which equals a dot product of the $\ell^2$-normalized descriptors.

\subsection{Results} 
For the evaluation, we split the dataset into a type-independent training and test set. 
For simplicity, we choose all images of the easy types (\#samples: 7029) as one subset and all the images of the difficult types (\#samples: 2043) as the other subset. 
We then evaluate the following two configurations: (1) trained with the difficult subset and tested with the easy one and (2) the other way around, \ie trained with the easy subset and tested with the difficult one.
We made sure that there is no type overlap in the subsets to guarantee type independent testing.
Note that this is different from the classification scenario, where we know the classes in advance.

We report typical retrieval measures, such as Top-1 accuracy as well as mean average
precision (\map), which is a measure of the overall ranking of the relevant
documents in respect to the query sample. 
Additionally, we give the Top-10 accuracy, \ie the chance of finding at least one sample of the query type among the first ten ranked results. 

\begin{table}[t]
	\centering
    \caption{Retrieval results for \subref{tab:easy} easy testset using the
		difficult for training and \subref{tab:difficult} difficult testset using
	the easy one for training. The first row denotes a document-dependent
scenario, where all samples of the test set are used for each query sample. For
the other experiments, retrieved samples from the same document are ignored.}
    \label{tab:retrieval}
    \subcaptionbox{\label{tab:easy}Easy}{
			\setlength{\tabcolsep}{4pt}
        \begin{tabular}{llccc}
            \toprule
            &Sampling & Top-1 & Top-10 & \map\\
            \midrule
            1vsAll&Keypoint+SIFT & 98.9 & 99.9 & 62.7\\
            \midrule
						\multirow{4}{*}{1vsOtherDocs} & Keypoint+SIFT       & 88.9 & 94.2 & 54.6\\
			&Keypoint+SIFT+ESVM  & 93.3 & 95.3 & 56.8\\
            &Contour+SIFT        & 72.8 & 80.8 & 47.1\\
            &Contour+SIFT+ESVM   & 76.5 & 80.2 & 49.2\\
            \bottomrule
        \end{tabular}
    }
    \subcaptionbox{\label{tab:difficult}Difficult}{
        \begin{tabular}{ccc}
            \toprule
            Top-1 & Top-10 & \map\\
            \midrule
						99.8 & 99.9 & 96.3\\
						\midrule
						50.0 & 50.2 & 46.6\\
						50.0 & 50,3 & 47.1\\						
						49.9 & 65.4 & 57.3\\
						49.9 & 65.4 & 57.3\\
            \bottomrule
        \end{tabular}
    }
\end{table}
First, we compute the typical leave-one-image-out scenario, \ie every test sample is used as query and \emph{all} remaining ones are ranked according to their similarity with the query.
\Cref{tab:retrieval} (first row) shows that this works astonishingly well with rates beyond \SI{99}{\percent}. 
This is only natural, since each query sample comes from a specific document, and the remaining document images are among the other samples of the test set. In other words, the algorithm most probably retrieves images from the same document. 

For the remaining experiments, we evaluate the retrieval performance in a document-independent way.
Therefore, we ignore images from the same document during the metric computation.
This results in a drop in performance but is a much more realistic scenario. 
When comparing the different sampling methods (keypoint vs.\ contour), we see
that keypoint-based sampling is superior to contour-based sampling when trained
on the difficult types samples and tested on the easy ones, see \cref{tab:easy}. 
The reverse behavior is shown for the difficult cases where contour-based
sampling is in favor, \cf \cref{tab:difficult}.  
This might be related to the quite imbalanced test document sizes; 
one of the difficult types has only one image in one document, thus the metrics are biased when retrieving this specific one.
This might also be the reason for the bad results for the difficult subset.  
ESVMs show also to be beneficial for type retrieval, at least for the easy cases. 

\section{Conclusion}
\label{sec:conclusion}
In this work, we analysed the possibilities of type recognition in the two scenarios classification and retrieval. 
Therefore, we adopted and evaluated two baseline systems, originally developed for font-group classification and writer identification.
While the classification network achieves an overall low accuracy, this can be attributed to the class imbalance of the training set.
In the case of type retrieval, a careful document-independent evaluation reveals that very similar-looking types are problematic for a common retrieval pipeline. 
For future work, we would like to investigate networks, trained by deep metric-learning methods, \eg with the use of contrastive or triplet loss, which might enable to differentiate also very similar-looking types.
\printbibliography

\end{document}